\PassOptionsToPackage{prologue,dvipsnames}{xcolor}
\documentclass[letterpaper, 10 pt, journal, twoside]{IEEEtran}

\usepackage{cite}
\usepackage{tikz}
\usepackage{graphicx}
\usepackage{amsfonts}
\usepackage{amsmath}
\usepackage[shortlabels]{enumitem}
\usepackage{subcaption}
\usepackage{booktabs}
\usepackage{colortbl}
\usepackage{arydshln}
\usepackage{url}
\usepackage[dvipsnames]{xcolor}
\usetikzlibrary{shapes.arrows}
\usepackage{hyperref}
\usepackage[T1]{fontenc}

\DeclareMathOperator{\sgn}{sgn}

\definecolor{red_color}{RGB}{255, 70, 70}
\definecolor{red_light_color}{RGB}{255, 210, 210}
\tikzset{
    RightUpArrow/.style = {
        single arrow,
        left color=red_light_color,
        right color=red_color,
        transform shape,
        minimum height=0.3cm,
        minimum width=0.2cm,
        inner sep=0.05cm,
        single arrow head extend=0.06cm,
        anchor=base,
        rotate=45,
        xshift=-5pt,
        yshift=-5pt,
    }
}
\setlist[enumerate]{label=\textbullet,
                    itemindent=0em,
                    itemsep=0.5em,
                    leftmargin=1em, % leftmargin=* if you like to have left margin as in main text
                    }

\title{A Champion-level Vision-based Reinforcement Learning Agent for Competitive Racing \\ in Gran Turismo 7
}

\author{Hojoon Lee$^*$$^{1}$, Takuma Seno$^*$$^{2}$, Jun Jet Tai$^*$$^{3}$, Kaushik Subramanian$^{4}$, \\ Kenta Kawamoto$^{2}$, Peter Stone$^{5,6}$, and Peter R. Wurman$^{5}$ % <-this % stops a space
\thanks{Manuscript received: December, 18, 2024; Revised March, 20, 2025; Accepted April, 9, 2025.}%Use only for final RAL version
\thanks{This paper was recommended for publication by Markus Vincze upon evaluation of the Associate Editor and Reviewers' comments.} %Use only for final RAL version
\thanks{$^*$These three authors contributed equally.}
\thanks{$^{1}$Hojoon Lee is with KAIST, Daejeon, South Korea.}
\thanks{$^{2}$Takuma Seno and Kenta Kawamoto are with Sony AI, Tokyo, Japan. {\tt\footnotesize takuma.seno@sony.com}}
\thanks{$^{3}$Jun Jet Tai is with Coventry University, Coventry, UK.}
\thanks{$^{4}$Kaushik Subramanian is with Sony AI, Zürich, Switzerland.}
\thanks{$^{5}$Peter Stone and Peter R. Wurman are with Sony AI, New York, USA.}
\thanks{$^{6}$Peter Stone is also with the University of Texas at Austin, USA.}
\thanks{Hojoon Lee and Jun Jet Tai have worked on this project for their internships at Sony AI, Tokyo, Japan.}
\thanks{Digital Object Identifier (DOI): 10.1109/LRA.2025.3560873}
}

{Lee \MakeLowercase{\textit{et al.}}: A Champion-level Vision-based Reinforcement Learning Agent for Competitive Racing in Gran Turismo 7}

\begin{document}

\maketitle
%\thispagestyle{empty}
%\pagestyle{empty}

%%%%%%%%%%%%%%%%%%%%%%%%%%%%%%%%%%%%%%%%%%%%%%%%%%%%%%%%%%%%%%%%%%%%%%%%%%%%%%%%
\begin{abstract}
Deep reinforcement learning has achieved superhuman racing performance in high-fidelity simulators like Gran Turismo 7 (GT7). It typically utilizes global features that require instrumentation external to a car, such as precise localization of agents and opponents, limiting real-world applicability. To address this limitation, we introduce a vision-based autonomous racing agent that relies solely on ego-centric camera views and onboard sensor data, eliminating the need for precise localization during inference. This agent employs an asymmetric actor-critic framework: the actor uses a recurrent neural network with the sensor data local to the car to retain track layouts and opponent positions, while the critic accesses the global features during training. 
Evaluated in GT7, our agent consistently outperforms GT7's built-drivers. To our knowledge, this work presents the first vision-based autonomous racing agent to demonstrate champion-level performance in competitive racing scenarios.
\end{abstract}


\IEEEpeerreviewmaketitle

%%%%%%%%%%%%%%%%%%%%%%%%%%%%%%%%%%%%%%%%%%%%%%%%%%%%%%%%%%%%%%%%%%%%%%%%%%%%%%%%

% Let's add this in the camera ready version
% To remain consistency with a camera-ready paper, what about just adding it for now?
\section*{Supplementary Videos} %The accompanied supplementary video demonstrates the agents' behavior across 3 distinct racing scenarios.
This paper is accompanied by a video of the performance: \url{https://youtu.be/cWqKbFsJcpo}
%as well as a video of the robustness evaluation:\\ 
%\url{https://youtu.be/cWqKbFsJcpo}
% Jet, Jet Jet Jet JEt JEt EJtawekjrklewjaklrjwaklefjrklwe

\section{INTRODUCTION}

\IEEEPARstart{A}{utonomous} racing demands self-driving vehicles to make split-second decisions at high speeds in dynamic, adversarial environments. Traditional control-based methods rely on modular pipelines for perception, planning, and control, requiring extensive hand-engineering. In contrast, deep reinforcement learning (RL) unifies these components, enabling end-to-end policy learning directly from sensor data. This approach has achieved superhuman performance in high-fidelity simulators like Gran Turismo 7 (GT7) through large-scale, distributed training \cite{wurman2022outracing}.

Despite these successes, transferring RL-based methods from simulation to real-world applications remains challenging. Current approaches rely on global features requiring external instrumentation, such as track geometry and opponent locations \cite{folkers2019controlling, fuchs2021super, remonda2021formula, song2021autonomous, wurman2022outracing, remonda2024simulation, xiao2024anycar}. Acquiring accurate real-time global features is challenging and introduces latency, which impedes rapid decision-making essential for racing \cite{kazerouni2022slam_survey, macario2022slam_comprehensive}. This reliance on global features restricts the practicality of deep RL approaches in real-world racing scenarios.

\begin{figure}[t]
\begin{center}
\vspace{1mm}
\includegraphics[width=0.92\linewidth]{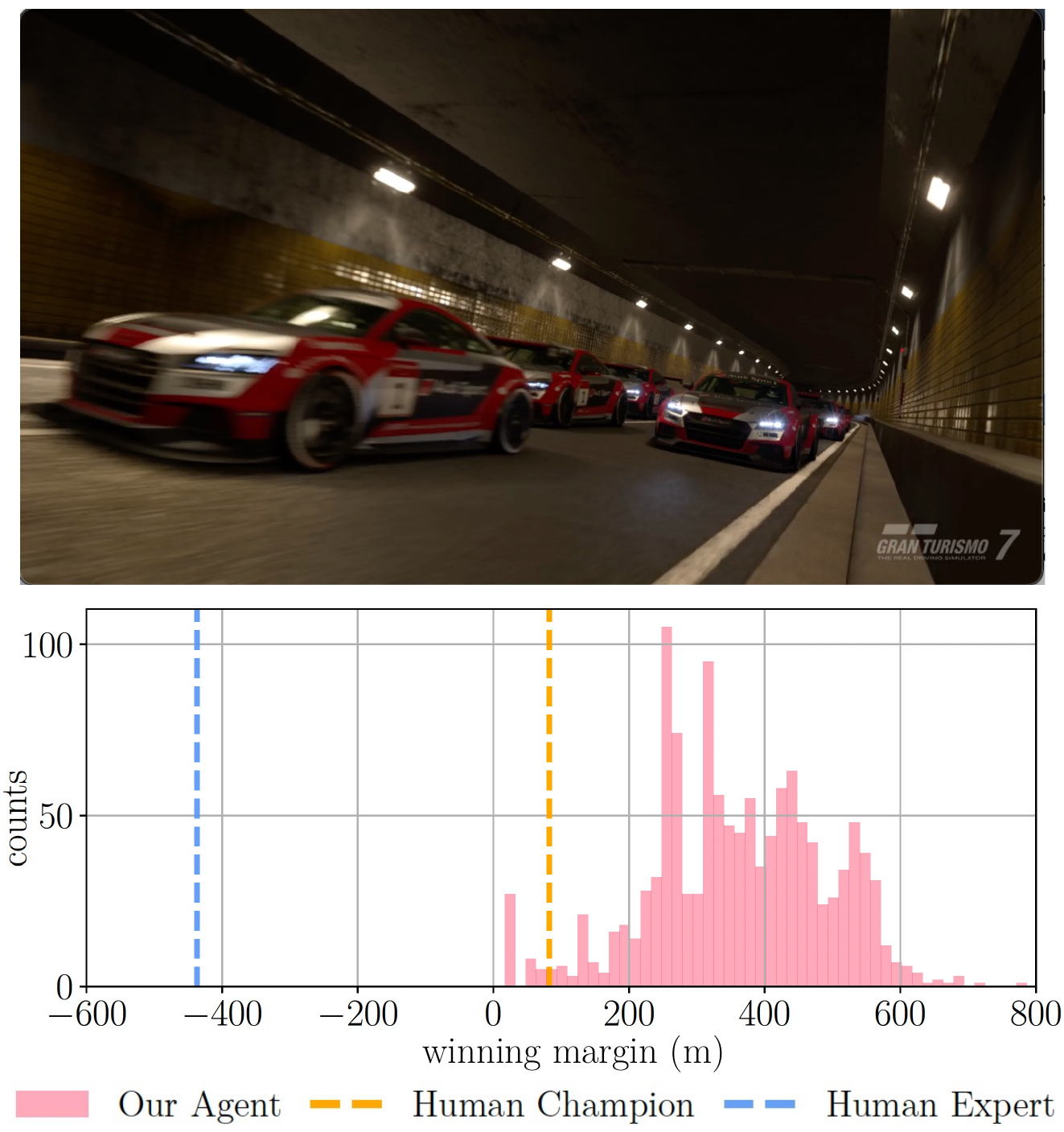}
\vspace{2mm}
\caption{\textbf{Top:} Our agent controlling an \textit{Audi TT Cup} and racing against GT7's built-in AI (BIAI). \textbf{Bottom:} Histogram of the \textit{winning margin}, the distance between our agent and the leading BIAI at race completion. This evaluation involves starting from the last position versus 19 identical BIAI agents on the \textit{Tokyo Expressway} track. Our agent consistently outperforms both Human Expert and Human Champion.}
\label{fig:intro}
\end{center}
\vspace{-3mm}
\end{figure}

A more feasible approach involves training agents using only onboard sensor data, such as ego-centric cameras and Inertial Measurement Units (IMUs), eliminating reliance on global features during inference. 
However using only on-board features can be challenging for competitive racing scenarios that include partial observability due to occlusions and track layouts, in addition to the difficulty of handling high dimensional image data\cite{hafner2019learning, yarats2021improving}.

Our work extends vision-based RL from time-trial \cite{vasco2024super} to competitive racing, where partial observability is more pronounced due to frequent occlusions of opponents and track layouts. We introduce an asymmetric recurrent actor-critic architecture \cite{pinto2017asymmetric, vasco2024super}, where the actor relies on vision-based input with a recurrent memory module \cite{kapturowski2018r2d2} to handle partial observability, while the critic leverages global state information during training. Additionally, to improve generalization and sample efficiency, we incorporate regularization techniques such as data augmentation \cite{laskin2020rad, yarats2021mastering} and periodic network reinitialization \cite{nikishin2022primacy, lee2024plastic}. Finally, we include a multi-opponent racing reward function from Wurman et al \cite{wurman2022outracing} alongside a reward function for the time-trial agent \cite{vasco2024super}. 

We evaluate our agent in GT7, a high-fidelity racing simulator for PlayStation® 5. Trained against GT7's built-in AI (BIAI), our agent consistently secures first place against 19 BIAIs, even when starting from the last position, outperforming human champions (Figure~\ref{fig:intro}). Extensive ablation studies validate the effectiveness of the asymmetric architecture, recurrent memory module, and regularization strategies. To the best of our knowledge, this work presents the first vision-based autonomous racing agent to achieve champion-level performance in competitive racing scenarios \cite{cai2021vision, herman2021learn, vasco2024super}.

%Although some studies have explored vision-based RL for autonomous racing, they often depend on costly expert demonstrations \cite{imamura2021expert, cai2021vision} or focus solely on time trial scenarios without opponents \cite{cai2021vision, chen2021safe, herman2021learn, vasco2024super}, limiting their applicability to competitive racing.

\section{RELATED WORK}
\subsection{Autonomous Racing}

Autonomous racing aims to develop vehicles capable of performing at their dynamic limits in competitive environments \cite{karle2022scenario, betz2022autonomous}. Traditionally, the problem has been divided into three main components: perception \cite{massa2020lidar, peng2021vehicle}, planning \cite{herrmann2020minimum, vazquez2020optimization}, and control \cite{williams2018robust, hao2022outracing, xue2024learning, kalaria2024adaptive, raji2022motion}, with progress often occurring in isolation. Recently, reinforcement learning (RL) has emerged as a powerful tool for integrating these components into unified, end-to-end systems \cite{fuchs2021super, remonda2021formula, song2021autonomous, wurman2022outracing, remonda2024simulation}. For example, Fuchs et al. \cite{fuchs2021super} achieved superhuman performance in a time trial racing scenario, where one car is on the track at a time, using a model-free RL approach with a novel reward structure. Subsequently, Wurman et al. \cite{wurman2022outracing} introduced Gran Turismo Sophy (GT Sophy), an RL agent that excels in both time trial races and multi-opponent races. However, during inference, these methods rely on global features, such as detailed track layout information and opponent localization, which are easily accessible in simulators but are challenging to obtain in real-world environments.

\subsection{Vision-Based RL for Autonomous Racing}

Vision-based RL presents a promising alternative by enabling agents to operate competitively directly from visual inputs, eliminating the need for precise global features during inference. Despite its potential, existing methods face significant challenges. Jaritz et al. \cite{jaritz2018end} reported that their vision-based RL agent struggled with maintaining optimal racing trajectories and frequently collided with obstacles. Cai et al. \cite{cai2021vision} combined imitation learning with model-based RL to teach racing behaviors, but their approach required costly expert demonstrations, limiting its scalability. Additionally, many vision-based methods either lack direct comparisons to human drivers \cite{remonda2021formula, jaritz2018end} or fail to perform effectively in competitive settings \cite{cai2021vision, herman2021learn}.

Vasco et al. \cite{vasco2024super} recently demonstrated that a vision-based agent could achieve superhuman performance in GT7. However, their work was confined to time trial settings without opponents, where partial observability and stochastic elements pose fewer challenges. In contrast, our work introduces the first vision-based RL agent to achieve champion-level performance in competitive racing scenarios, where the agent needs to interact with opponent cars and aim for the first position while respecting the rules of sportsmanship.

\section{METHOD}

%Previous implementations of superhuman autonomous racing RL agents in GT7 have relied on global features, such as low-dimensional state data that represent course points and opponent states \cite{wurman2022outracing}. 
Our goal is to develop a vision-based agent for competitive racing scenarios in GT7 using only sensor data local to the car during inference.
Our agent is built on top of the previous vision-based racing agent for time trial settings in GT7 \cite{vasco2024super}.

\subsection{Observation Space}

Our agent employs a multimodal observation space designed to capture the critical aspects of competitive racing. 
At each time step $t$, the composite observation $\mathbf{o}_t = (\mathbf{o}_t^i, \mathbf{o}_t^p, \mathbf{o}_t^g)$ consists of image data $\mathbf{o}_t^i$, proprioceptive information $\mathbf{o}_t^p$, and global information $\mathbf{o}_t^g$ derived from the GT7 simulation during training.

%\begin{enumerate}[itemsep=0.5em, left=0.0em, label=\textbullet]
\begin{enumerate}
    \item \textbf{Image Feature} (\( \mathbf{o}_t^i \)) is a \(64 \times 64\) RGB image, down-scaled from the original \(1920 \times 1080\) resolution. It captures the agent's first-person view of the track. To simulate a front-view camera attached to the car, we disabled the in-game heads-up-display containing information about the vehicle speed or track map and masked out the rear-view mirror. Note that Vasco et al. \cite{vasco2024super} showed that the rear-view mirror was not critical for race car control.
    %Additionally, we masked out the rear-view mirror. Since our agent primarily passes other BIAI vehicles rather than being passed, the rear-view mirror is not critical for its decision-making.
    %Some camera view options in GT7 don't provide the rear-view mirror in the screen. For fair comparison with human players, we masked out the rear-view mirror.
    
    \item \textbf{Proprioceptive Feature} (\( \mathbf{o}_t^p \))
    includes data from the IMU sensors, defined as:
    $$
    \mathbf{o}_t^p = \left[ 
        \mathbf{v}_t, \, 
        \mathbf{\dot{v}}_t, \, 
        \mathbf{v}_t^r, \, 
        \mathbf{u}_t, \, 
        \mathbf{s}_t, \, 
        \mathbf{d}_t 
    \right]
    $$
    where \( \mathbf{v}_t \in \mathbb{R}^3 \) represents the car's linear velocity, \( \mathbf{\dot{v}}_t \in \mathbb{R}^3 \) is the linear acceleration, and \( \mathbf{v}_t^r \in \mathbb{R}^3 \) is the rotational velocity. The vector \( \mathbf{u}_t \in \mathbb{R}^3 \) corresponds to the current inputs for steering, throttle, and brake. Lastly, \( \mathbf{s}_t \in \mathbb{R}^3 \) and \( \mathbf{d}_t \in \mathbb{R}^3 \) are the steering angle and changes in the steering angle over the last three time steps.
    
    \item \textbf{Global Feature} (\( \mathbf{o}_t^g \))  includes track point information, $c_t$, and opponent grid data, $g_t$, proposed in Wurman et al. \cite{wurman2022outracing}.
     The track point feature consists of 177 3D coordinates \( \mathbf{c}_t \in \mathbb{R}^{177 \times 3} \), representing the edge of the track. These points are dynamically spaced based on the agent's speed, covering approximately six seconds of travel time.
    
    The opponent grid feature contains information about nearby opponents. For each opponent, the position, velocity, and acceleration, projected onto the 2D plane, are recorded, forming a vector \( \mathbf{g}_t^{\text{opp}} \in \mathbb{R}^6 \). The full opponent grid feature is represented as \( \mathbf{g}_t \in \mathbb{R}^{6 \times 14} \), describing the 7 closest opponents looking ahead 75 meters ahead and 7 closest opponents looking behind 20 meters.
\end{enumerate}

While the critic uses global features during training, they are excluded from the actor to ensure that the agent relies exclusively on information local to the car during inference.

\subsection{Action Space}

We follow the action space used in previous work \cite{vasco2024super}:
$$
\mathbf{a}_t = (a_t^s, a_t^g)
$$
where \( a_t^s \in \mathbb{R} \) represents the \textit{delta steering angle}, constrained to the range \([-3^\circ, 3^\circ]\) to ensure realistic steering inputs. The term \( a_t^g \) denotes the \textit{combined throttle and brake} value, within the range \([-1, 1]\), with \(-1\) indicating full braking and \(1\) indicating full throttle. Gear shifting is managed by the in-game automatic transmission system.

The agent's control updates occur at a frequency of 10 Hz, whereas the game operates at 60 Hz. To synchronize, the game applies a zero-order hold for throttle inputs and linearly interpolates the steering angle between control updates.

\subsection{Reward Function}

We utilize a reward function used in previous work \cite{vasco2024super}, incorporating a multi-opponent racing reward function from Wurman et al. \cite{wurman2022outracing}, defined as a weighted combination of atomic reward components:
\begin{align*}
    \begin{split}
        r_t = \lambda^p r_t^p &+ \lambda^o r_t^o + \lambda^b r_t^b + \lambda^v r_t^v \\
        &+ \lambda^c r_t^c + \lambda^s r_t^s + \lambda^t r_t^t + \lambda^h r_t^h.
    \end{split}
\end{align*}

%\begin{enumerate}[itemsep=0.5em, left=0.0em, label=\textbullet]
\begin{enumerate}
    \item  \textbf{Track Progress} ($r^p$) measures the one-step change in the vehicle's track position since the last step. It is defined as \( r_t^p = p_t - p_{t-1} \), where \( p_t \) represents the vehicle's position projected onto the closest point on the track center line.

    \item \textbf{Shortcut Penalty} ($r^o$) penalizes the agent for taking shortcuts by cutting track corners. It is defined as \( r_t^o = - (s_t^o - s_{t-1}^o) |\mathbf{v}_t| \), where \( s_t^o \) denotes the total time the vehicle has had at least three tires outside the track limits.

    \item \textbf{Barrier Collision Penalty} ($r^b$) discourages the agent from using barrier collisions to change directions quickly. It is defined as \( r_t^b = - (s_t^b - s_{t-1}^b) |\mathbf{v}_t| \), where \( s_t^b \) represents the total time the vehicle was in contact with a barrier.

    \item \textbf{Car Velocity-based Collision Penalty} ($r^v$) penalizes the agent for colliding with other cars based on the difference in speed. It is defined as \( r_t^v = - |\Delta \mathbf{v}_t^x|^2 \), where \( \Delta \mathbf{v}_t^x \) is the speed difference between the agent's car and an opponent's car at the time of collision

    \item \textbf{Car Collision Fixed Penalty} ($r^c$) applies a constant penalty for any contact with opponent cars.

    \item \textbf{Overtaking Progress} ($r^t$) rewards the agent for overtaking other cars. It is defined as \( r^t = \sum_{\forall i \in C \backslash k} [ \mathbb{I}_{c_r < (p_t^i - p_t) < c_f} ((p_t - p_t^i) - (p_{t-1} - p_{t-1}^i)) ] \), where \( p_t^i \) represents the position of car \( i \), \( k \) is the index of the ego-agent's vehicle, and $c_r$ and $c_f$ are thresholds for the minimum distance between two cars.

    \item \textbf{Steering Change Penalty} ($r^s$) discourages abrupt steering changes. It is defined as \( r_t^s = - |\theta_t^s - \theta_{t-1}^s| \), where \( \theta_t^s \) is the steering angle at time \( t \).

    \item \textbf{Steering History Penalty} ($r^h$) penalizes inconsistent steering decisions over a short period. It is defined as \( r_t^h = -m_t (1 + \exp(-c^s \cdot (\Delta_t - c^o))) \), where \( \Delta_t = |\delta_t| + |\delta_{t-1}| \), \( \delta_t = \theta_t^s - \theta_{t-1}^s \) and \( m_t = \mathbb{I}_{\delta_t > c^d} \cdot \mathbb{I}_{\delta_{t - 1} > c^d} \cdot \mathbb{I}_{\sgn(\delta_t) \neq \sgn(\delta_{t-1})} \). $c^s$, $c^o$, and $c^d$ are constant factors.
\end{enumerate}

Following previous work \cite{wurman2022outracing, vasco2024super}, we used $\lambda$ values as: \( \lambda^p = 1.0 \), \( \lambda^o = 10.0 \), \( \lambda^b = 20.0 \), \( \lambda^v = 0.5 \), \( \lambda^c = 6.0 \), \( \lambda^s = 0.5 \), \( \lambda^t = 3.0 \), and \( \lambda^h = 5.0 \). The constant factors are $c_r=-20$, $c_f=40$, $c^s=182.883569$, $c^o=0.034$, and $c^d=0.014$.

\begin{figure}[t]
\begin{center}
\vspace{1mm}
\includegraphics[width=0.85\linewidth]{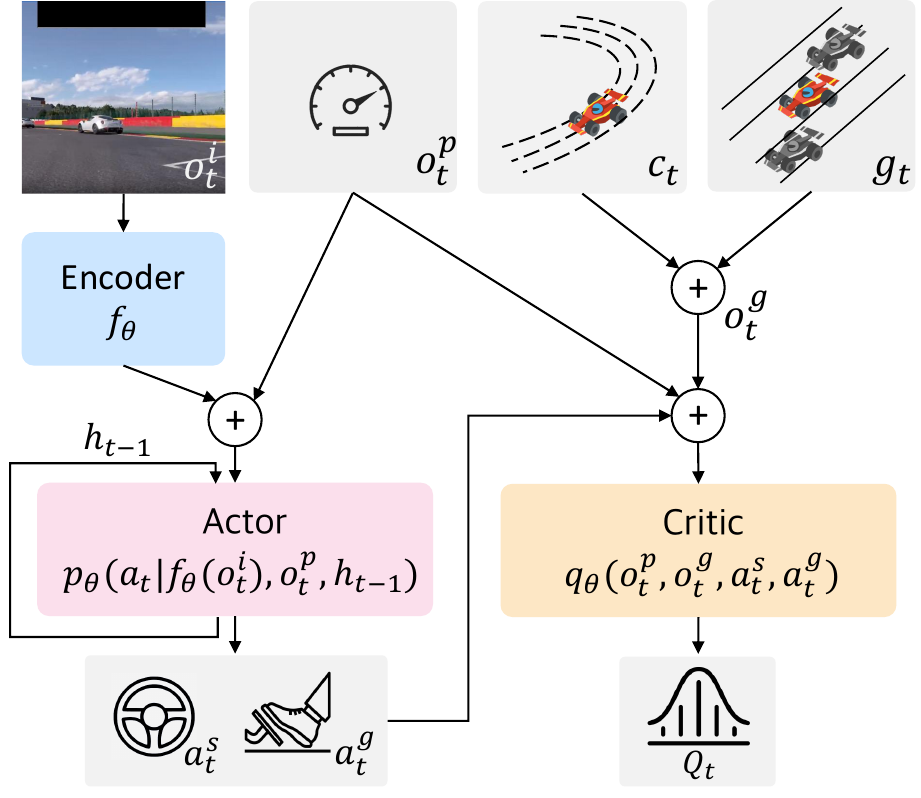}
\vspace{1mm}
\caption{\textbf{Architecture Overview.} The actor processes the image and proprioceptive features to predict actions, using a recurrent memory to track opponents and track layouts. The critic evaluates these actions using the global features. Both networks are jointly trained with the QR-SAC algorithm.} \label{fig:architecture}
\vspace{-3mm}
\end{center}
\end{figure}

\begin{table*}[t]
    \centering
       \caption{We evaluate our agents across three distinct scenarios, each consisting of a track, car, and tire combination.}
    \begin{tabular}{llll}
    \toprule
        Scenario & Track & Car & Tire \\
    \midrule
         \arrayrulecolor{gray!20}\hdashline\arrayrulecolor{black}
        {\ttfamily Tokyo} & Tokyo Expressway - Central Clockwise, Japan & Audi TT Cup '16 & Racing Hard \\
        {\ttfamily Spa} & Circuit de Spa-Francorchamps, Belgium & Alfa Romeo 4C Launch Edition ’14 & Sports Medium \\
         \arrayrulecolor{gray!20}\hdashline\arrayrulecolor{black}
        {\ttfamily Sarthe} & 24 Heures du Mans race track, France & HYUNDAI N 2025 Vision Gran Turismo (Gr.1) & Racing Medium \\
    \bottomrule
    \end{tabular}
    \label{tab:scenarios}
\end{table*}

\subsection{Architecture}

We use Quantile Regression Soft Actor-Critic (QR-SAC) \cite{dabney2018distributional}, a distributional variant of SAC \cite{haarnoja2018soft} to train the agent.
This algorithm has successfully learned superhuman autonomous racing agents in previous work \cite{wurman2022outracing, vasco2024super}.

As illustrated in Figure \ref{fig:architecture}, we use an \textit{asymmetric} actor-critic architecture, which is designed as follows:

%\begin{enumerate}[itemsep=0.5em, left=0.0em, label=\textbullet]
\begin{enumerate}
    \item \textbf{Actor} $(\pi_{\theta})$: The actor is only provided with the local features, the image input \( \mathbf{o}_t^i \) and proprioceptive data \( \mathbf{o}_t^p \). 
    The image is passed through three convolutional layers, $f_{\theta}$, They use 32, 64, and 64 filters respectively. The kernel sizes are 8, 4, and 3 with strides of 4, 2, and 1. The resulting feature map is flattened and embedded into a 512-dimensional vector, which is then concatenated with the proprioceptive features. This combined feature vector is then processed by a recurrent predictor network, $p_{\theta}$, which includes an internal hidden state, $h_{t-1}$. The recurrent module is implemented with a Gated Recurrent Unit \cite{cho2020learning}, followed by four fully connected layers, each with 2048 hidden units. 
    A final linear layer with a hyperbolic tangent activation function predicts action probabilities of the \textit{delta steering angle} and the \textit{combined throttle  and brake} value individually, modeled by a Gaussian distribution.
    
    \item \textbf{Critic} $(q_{\theta})$: The critic uses both proprioceptive data \( \mathbf{o}_t^p \) and global features \( \mathbf{o}_t^g \) to precisely evaluate actions based on local and global information.
    The network consists of 4 fully connected layers with 2048 hidden units each and outputs a value function with 32 quantile units to model the Q-function distribution.

\end{enumerate}

Using this asymmetric architecture, the actor relies on image and proprioceptive features, allowing the agent to make inferences based solely on local information.

\subsection{Regularization}

To improve the stability and generalization of our vision-based agent, we apply the following regularizations:

%\begin{enumerate}[itemsep=0.5em, left=0.0em, label=\textbullet]
\begin{enumerate}

\item \textbf{Network Reinitialization}: In RL, agents can overfit to early training data which often includes limited behaviors, such as navigating simpler track sections or less dynamic opponent interactions \cite{lee2024plastic}. This can lead to overemphasis on static features like track layouts. To alleviate this bias, we reinitialize the networks after the replay buffer is fully populated, as recommended by Nikishin et al. \cite{nikishin2022primacy}. At this stage, the buffer contains a diverse range of scenarios, including complex opponent interactions and strategic behaviors. Reinitializing the network allows the agent to restart from this broader dataset and helps to prevent the agent from prematurely overfitting to static features.

% \item \textbf{Network Reinitialization}: In RL, the agent can overfit to early interactions as the replay buffer is initially populated with transitions from the early stages of training.
% This may cause the agent to focus on features like track layouts while neglecting opponent features. To address this, we reinitialize the network once after filling the buffer \cite{nikishin2022primacy}, enabling the agent to relearn features from diverse samples.

% \item \textbf{Alpha Scheduling}: The temperature parameter $\alpha$ in QR-SAC, which determines the trade-off between exploration and exploitation, is dynamically scheduled. A cosine decay schedule gradually reduces $\alpha$, allowing the agent to explore effectively during early training while converging to stable policies in later stages.

\item \textbf{Image Augmentation}: To prevent overfitting to specific visual cues, we apply random shift augmentation \cite{kostrikov2020image}. During training, the input image is randomly shifted within a small range, simulating different visual perspectives and enhancing the agent’s ability to generalize to unseen scenarios.

%\item \textbf{RNN Burn-in}: Inspired by the R2D2 architecture \cite{kapturowski2018r2d2}, the recurrent network in the actor is initialized with a burn-in phase of 16 steps during training. This ensures that the recurrent states are preconditioned to capture temporal dependencies effectively, especially for long-horizon tasks such as autonomous racing.

\end{enumerate}

These regularization strategies improve the agent’s learning stability and generalization, supporting robust decision-making in competitive racing environments.

\begin{figure}[t]
\begin{center}
\includegraphics[width=0.98\linewidth]{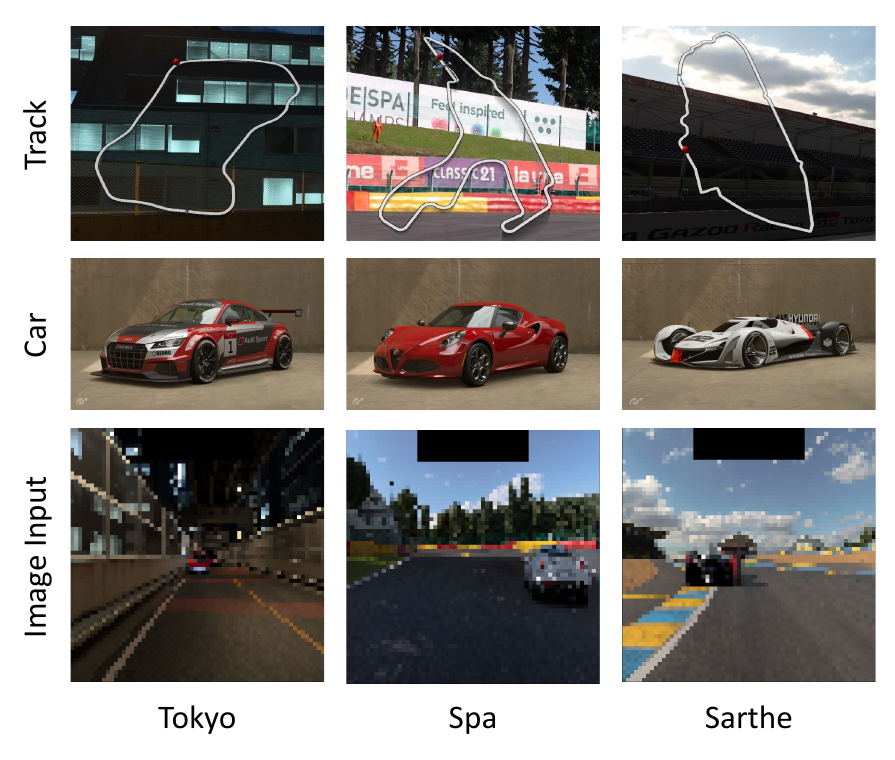}
\end{center}
\vspace{-2mm}
\caption{\textbf{Racing Scenarios.} Visualization of track, car, and sample input image for each training scenario.}
\vspace{-2mm}
\label{figure:scenario}
\end{figure}

\section{EXPERIMENTAL SETUP}
 
\subsection{Environments}

We evaluate our approach in GT7 across three car-track scenarios, each presenting different challenges. Detailed setups are provided in Table \ref{tab:scenarios} and Figure \ref{figure:scenario}.

%\begin{enumerate}[itemsep=0.5em, left=0.0em, label=\textbullet]
\begin{enumerate}
    \item  \textbf{Tokyo:} A track with a mix of chicanes, high-speed straights, and tight track boundaries with no run-off, requiring precise control in overtaking maneuvers with a front-wheel drive vehicle.
    \item \textbf{Spa:} A technical circuit with significant elevation changes, demanding good racing lines and control of vehicle oversteer with a rear-wheel drive vehicle. 
    \item \textbf{Sarthe:} A high-speed circuit where effective slipstreaming on long straights and managing vehicle downforce are critical, with speeds reaching 340 km/h in a 4WD car. 
\end{enumerate}

\begin{figure*}
\begin{center}
\begin{subfigure}[b]{0.32\linewidth}
    \includegraphics[width=\linewidth]{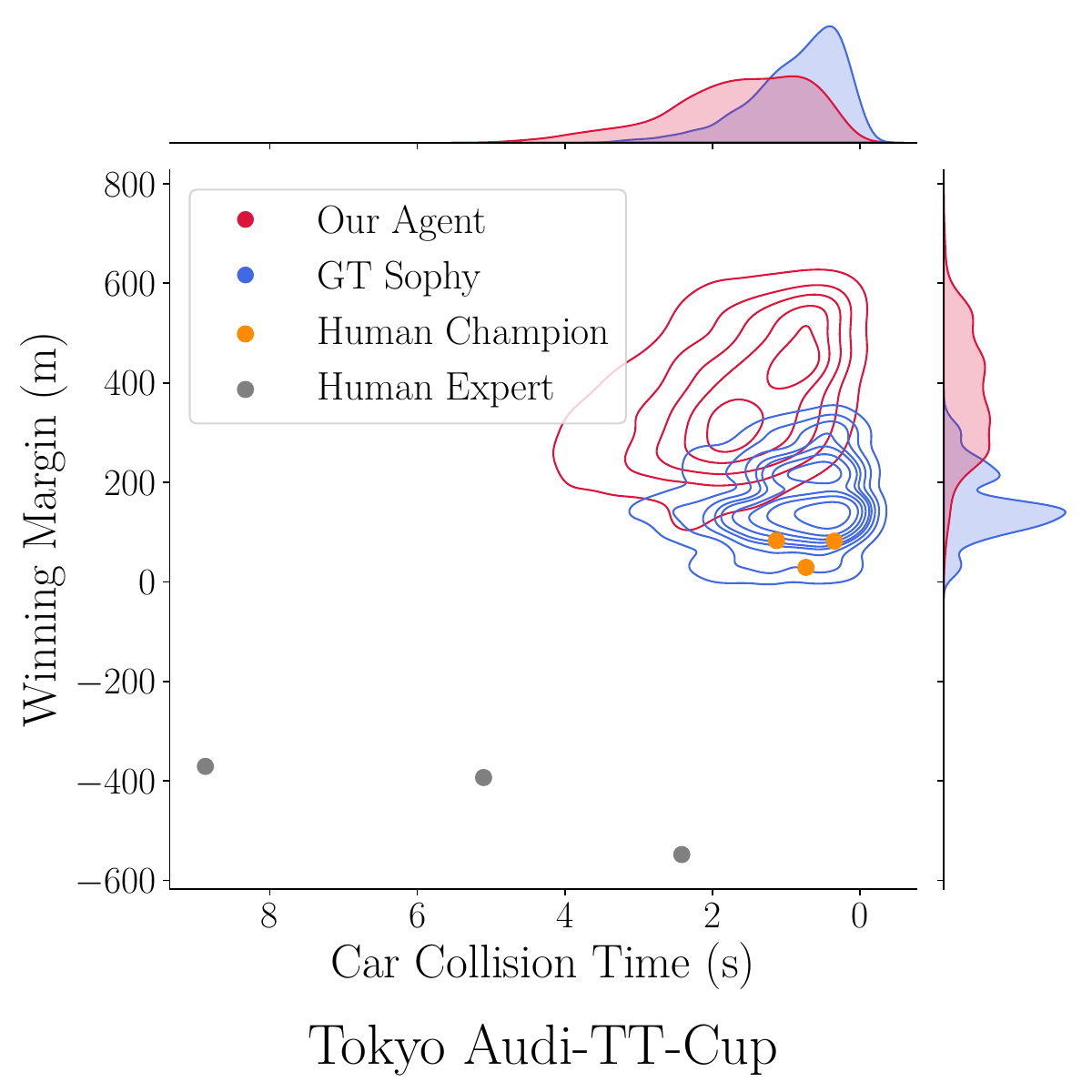}
    %\caption{Tokyo Audi-TT-Cup}
    \label{fig:tokyo_audi}
    \phantomcaption{}
\end{subfigure}
\begin{subfigure}[b]{0.32\linewidth}
    \includegraphics[width=\linewidth]{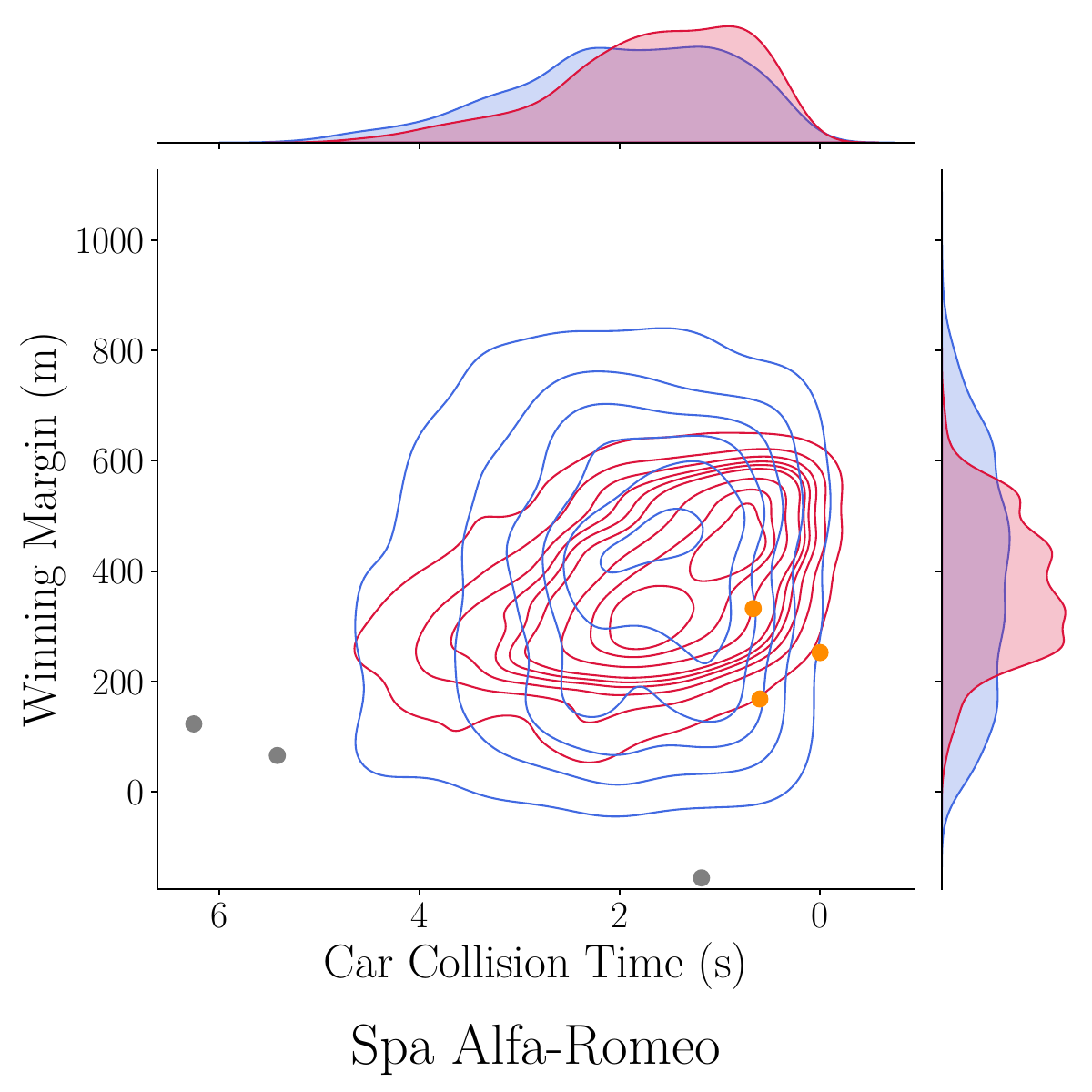}
    %\caption{Spa Alfa-Romeo}
    \label{fig:spa_alfa}
    \phantomcaption{}
\end{subfigure}
\begin{subfigure}[b]{0.32\linewidth}
    \includegraphics[width=\linewidth]{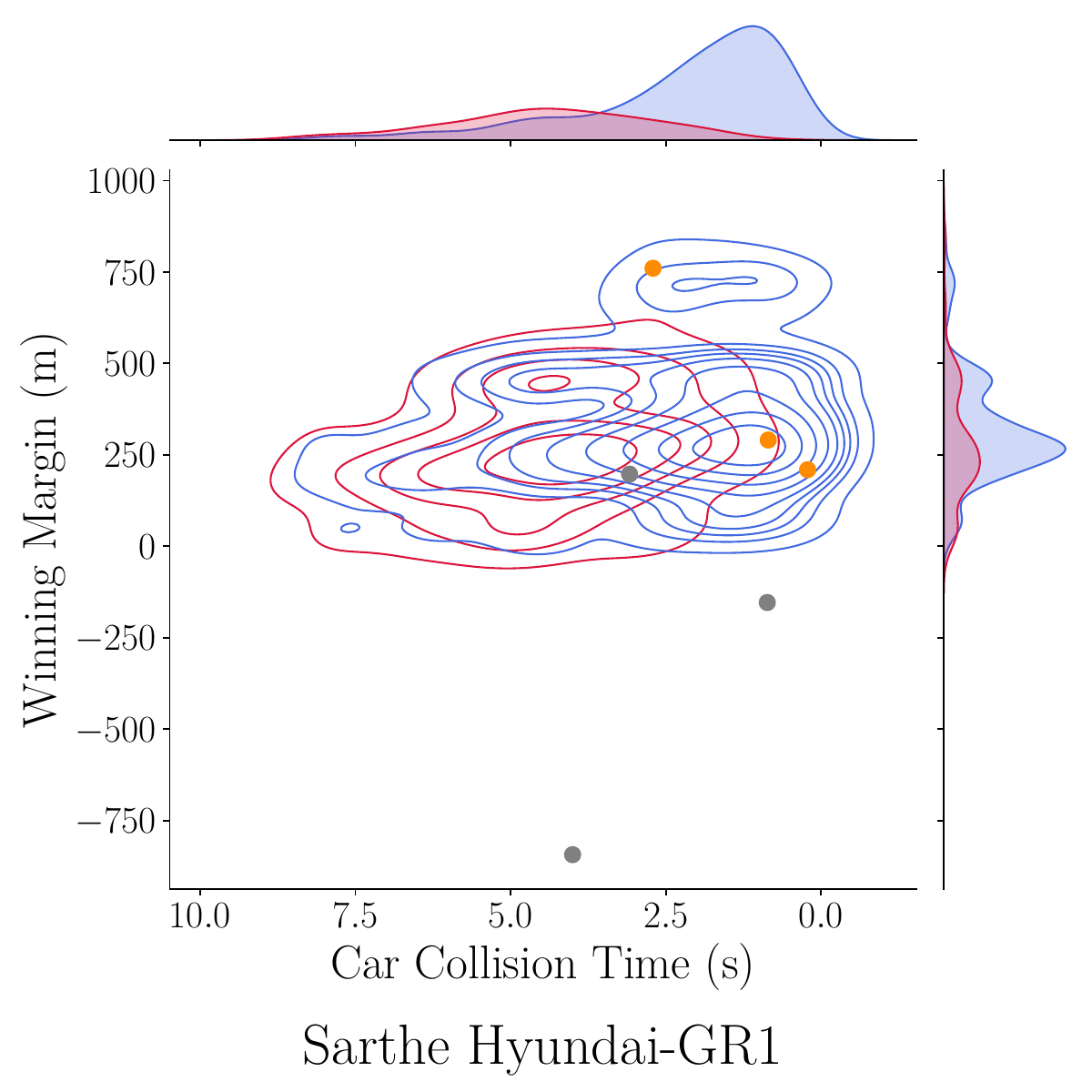}
    %\caption{Sarthe Genesis-GR1}
    \label{fig:sarthe_hyundai}
    \phantomcaption{}
\end{subfigure}
\vspace{-4mm}
\caption{\textbf{Performance comparison of our agent, GT Sophy, a Human Expert, and a Human Champion.} \textit{Car collision time} is the total duration of contact with any opponent, while \textit{winning margin} is the distance between the agent and the highest-ranked opponent when the agent completes all four laps. The contours represent the density of data points, with denser regions indicating more frequent occurrences of certain performance outcomes. Upper-right regions indicate superior performance, as they represent larger \textit{winning margin} achieved with lower \textit{car collision times}.}
\vspace{-2mm}
\label{fig:baseline_results}
\end{center}
\end{figure*}

\subsection{Training}

Unlike many RL simulators (e.g., MuJoCo \cite{todorov2012mujoco}) that can leverage accelerated simulation, GT7 operates in real-time. To speed up training, we used the same asynchronous distributed training framework described in Wurman et al. \cite{wurman2022outracing}. Our setup utilized 20 rollout workers for data collection, each connected via Ethernet to a dedicated PlayStation® 4 system.

Latency from retrieving images over Ethernet presented challenges for real-time training. To mitigate this problem, we configured the simulator to pause simulation steps until action commands were received from the rollout workers \cite{vasco2024super}. A dedicated training server managed the network parameters and updated them via gradient descent. Rollout workers are synchronized with the server at the end of each epoch by receiving the latest policy checkpoint.

While GT7 supports a maximum of 20 cars per track, training against 19 BIAI opponents directly can hinder the agent's ability to learn basic driving skills. 
To address this challenge, we adopted the multi-scenario training approach from Wurman et al. \cite{wurman2022outracing}. Each training episode is sampled from a range of configurations, from solo runs to races with 1, 2, 3, 4, 7, 12, and 19 opponents.
The agent starts each race from randomly sampled points around the track.
%with some configurations featuring the agent surrounded by 19 opponents to help it learn to navigate crowded situations. 
To diversify opponent behavior, GT7's balance-of-performance (BoP) interface uniformly samples the opponent cars' engine power and body weight within [-25\%, + 25\%] range relative to vehicle's original specification.

For the main experiments,  we adopted the same hyperparameters as those established in prior work \cite{vasco2024super}. We used a mini-batch size of 512, sampled from 16 trajectories with a sequence length of 32 and a burn-in phase of 16 steps for the recurrent module. The replay buffer was set to 5 million samples, and we applied the Adam optimizer with a learning rate of $2.5 \times 10^{-5}$, a discount factor of 0.9896, and an entropy coefficient of 0.01. We incorporated a multi-step return with $n=7$.
To encourage the network to relearn overlooked features, we reinitialized the networks at 2,000 epochs, which aligned with the replay buffer reaching capacity. Image inputs were augmented using random shift with a maximum shift of 4 pixels, utilizing mirrored padding. The training was conducted over 20,000 epochs.

\subsection{Baselines}

For comparison, we include the following baselines: 
%to evaluate the performance of our agent:

\vspace{2mm}
%\begin{enumerate}[itemsep=0.5em, left=0.0em, label=\textbullet]
\begin{enumerate}

\item \textbf{Human Expert:}\footnote{We conducted evaluation trials with Rodney Meza to present results as a Human Expert. He is an employee of Sony Research, Tokyo.} Performance of a GT7 player with 25+ years of experience in the Gran Turismo series and real-world circuit racing. The player is regularly ranked in the top 3-5\% in online time trial events of GT7. The player was allowed unlimited practice laps and evaluated in three trials per scenario, with trials restarted if the player lost control and spun out to ensure consistency. 
\item \textbf{Human Champion:}\footnote{We conducted evaluation trials with Mikail Hizal, a champion at the GT World Series events in 2019 and 2020, to present results as a Human Champion.} Performance of a top GT7 player with multiple world titles. The evaluation protocol matched that of the Human Expert.
\item \textbf{GT Sophy:} An agent trained using the architecture described in Wurman et al. \cite{wurman2022outracing}, with modifications to train exclusively against BIAI. Population-based training is not utilized in this setup.
%\item \textbf{Our Agent:} Our proposed agent, which relies solely on visual and proprioceptive inputs during inference, without access to any global information.

\end{enumerate}

\subsection{Evaluation}

We evaluated the performance of each agent over 4-lap episodes in their respective scenarios, starting from the back of a 20-car grid, with the remaining 19 cars controlled by BIAI. 
BIAI is an in-game model predictive control-based AI in GT7, serving as the opponent for all experiments. 
The BoP is disabled during evaluation. The primary metric was the \textit{winning margin}, which measures the distance by which the agent leads the second-best opponent if it wins, or the deficit if it does not. Sportsmanship was also considered, with overtaking through collisions discouraged. We tracked \textit{car collision time}, the total duration the agent’s car was in contact with others during an episode.

After training, we selected the top model checkpoint for each seed (three seeds in total) based on the highest \textit{winning margin} with a \textit{car collision time} better than the worst observed in the human champion baseline. The evaluation was conducted over 500 episodes per model checkpoint.

\begin{figure*}
\begin{center}
\includegraphics[width=0.85\linewidth]{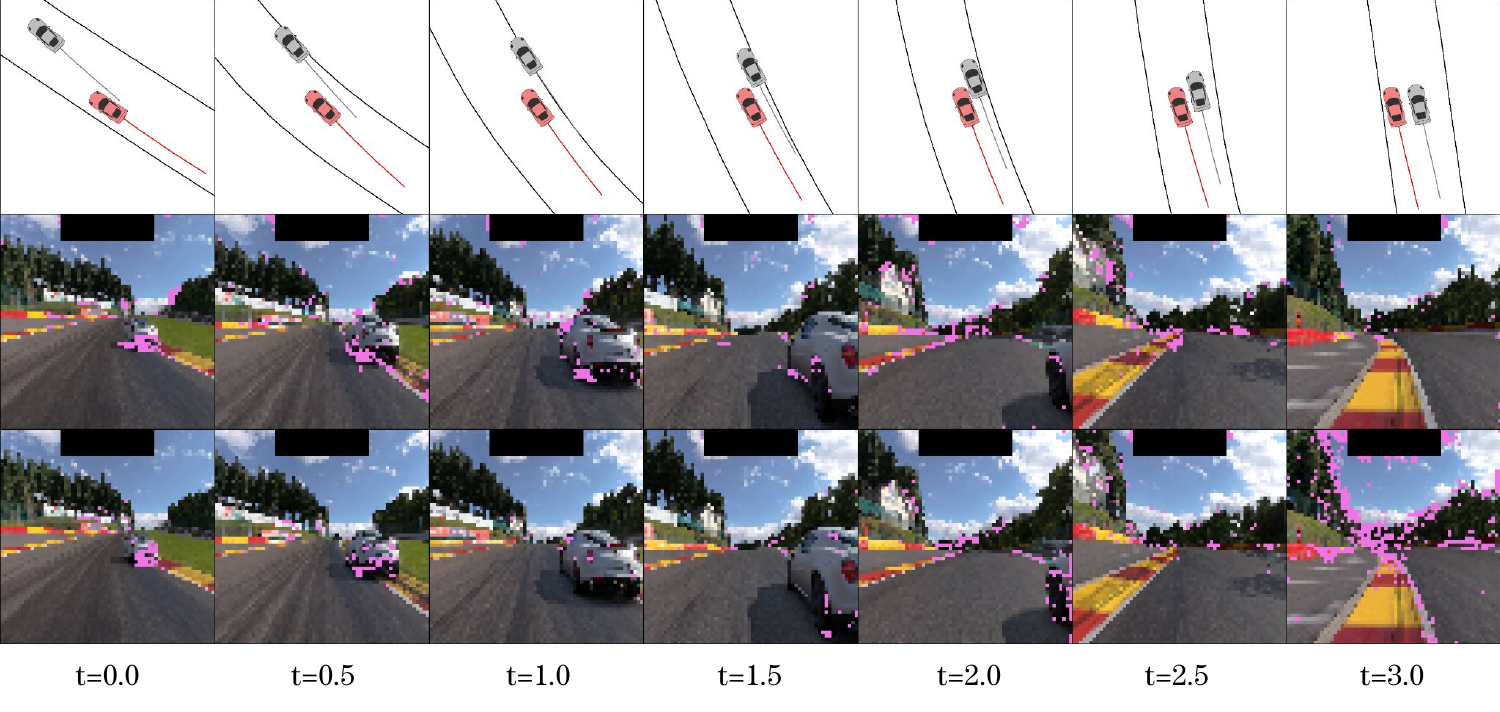} 
\caption{\textbf{Visualizing our agent's trajectory and action attributions in the Spa scenario.} The sequence is shown in 0.5-second intervals and consists of three rows: \textbf{Top:} displays the trajectory of our agent (red) and a BIAI opponent (black); \textbf{Middle:} shows attribution maps using Integrated Gradients, highlighting the agent's focus on lower vehicle regions for overtaking opportunities or treelines for track layout. \textbf{Bottom:} illustrates how visual features from the past frames contribute to actions predicted for the final frame, demonstrating the agent's ability to infer information that is not included in the final frame.}
\label{fig:integrated_gradients}
\end{center}
\vspace{-2mm}
\end{figure*}

\section{EXPERIMENTAL RESULTS}

\subsection{Main Experiment}

We present the main experiment results in Figure \ref{fig:baseline_results}, using a Kernel Density Estimate plot \cite{Scott1992} to visualize the relationship between \textit{car collision time} (x-axis) and winning margin at the final distance (y-axis). We focus on the 50\% interquartile range, with better performance indicated by density concentrated on the upper-right corner (\begin{tikzpicture}\node [RightUpArrow] at (0,0) {};\end{tikzpicture}).
In all scenarios, our agent demonstrates champion-level performance:
\begin{itemize}
    \item \textbf{Tokyo.} Our vision-based agent outperforms all baselines, achieving the highest \textit{winning margin}. This superior performance likely stems from our agent's ability to assess the distance and gaps to nearby cars better than GT Sophy. While GT Sophy treats opponents as point masses with relative position, velocity, and acceleration, it lacks awareness of opponent orientation \cite{wurman2022outracing}. In contrast, our agent’s vision-based input enables it to infer opponent orientation, leading to improved gap perception and overtaking capabilities. These advantages are particularly critical on the Tokyo track, which demands precise navigation within tight boundaries and minimal run-off areas.
    
    \item \textbf{Spa.} Our agent matches GT Sophy’s performance while surpassing the human expert and the human champion baseline. Overtaking is easier here due to the track's wider layout and generous run-off areas, allowing our agent to exploit curbs and optimize its racing line. The similar performance between our agent and GT Sophy is expected, as both consistently execute near-perfect racing lines, providing a significant advantage over the human champion, whose racing lines are skilled but less precise.

    %\item \textbf{Sarthe.} On Sarthe, our agent induces higher \textit{car collision time} than the human champion. Although we selected a model checkpoint with a collision time lower than the worst observed human performance, the agent’s overall collision time remains higher. This discrepancy is likely due to inherent randomness in GT7: even with identical starting conditions, vehicle positions can diverge significantly after the first corner, where a majority of collisions occur. Nevertheless, our vision-based agent consistently surpasses the human expert baseline in performance, maintaining champion-level results.

    \item \textbf{Sarthe.} On Sarthe, our agent consistently surpasses the human expert and a majority of the human champion data in performance. Note that our agent induces higher \textit{car collision time} compared to the human champion. Although we selected a model checkpoint with a collision time lower than the worst observed human performance, the agent’s overall collision time remains higher. This discrepancy is likely due to inherent randomness in GT7: even with identical starting conditions, vehicle positions can diverge significantly after the first corner, where a majority of collisions occur. GT Sophy achieves a lower collision time than our agent, utilizing the precise perception to avoid collisions at the first corner.
    
    %\item \textbf{Sarthe} On Sarthe, our agent struggles to match the human champion’s \textit{car collision time}. Although we selected a model checkpoint with a collision time lower than the worst observed human performance, the agent’s overall collision time remains higher. We closely checked evaluation episodes and found out that most collisions occur near the first corner, where the agent's starting position varies between runs. We speculate that the vision-based agent may struggle to generalize to unfamiliar visual inputs not seen during training. Despite this, our vision-based agent consistently outperforms the human expert baseline, maintaining champion-level performance.

    % On Sarthe, our agent struggles to match the human champion’s \textit{car collision time}. Although we selected a model checkpoint with a collision time lower than the worst observed human performance, the agent’s overall collision time remains higher. This discrepancy is likely due to inherent randomness in GT7: even with identical starting conditions, vehicle positions can diverge significantly after the first corner, where a majority of collisions occur. Nevertheless, our vision-based agent consistently surpasses the human expert baseline in performance, maintaining champion-level results.
\end{itemize}

%In the Spa scenario, our agent matches the competitiveness of GT Sophy while outperforming the human champion baseline. Overtaking is considerably easier at Spa due to the track's wider layout and generous run-off areas, which allow our agent to exploit curbs and optimize racing lines for more efficient passes. In this context, the similar performance between our agent and GT Sophy is understandable --- both consistently execute near-perfect racing lines. This key distinction highlights their shared advantage over the human champion, whose racing lines and execution, while skilled, lack the precision achievable by machine models.

% Despite these challenges, our agent can still be considered to demonstrate champion-level performance on Sarthe. Once it attains first place (typically after 2 laps), the agent consistently takes better racing lines, which leads to improved \textit{winning margin last distance}s compared to the human champion.

\begin{figure*}
\begin{center}
\includegraphics[width=0.87\linewidth]{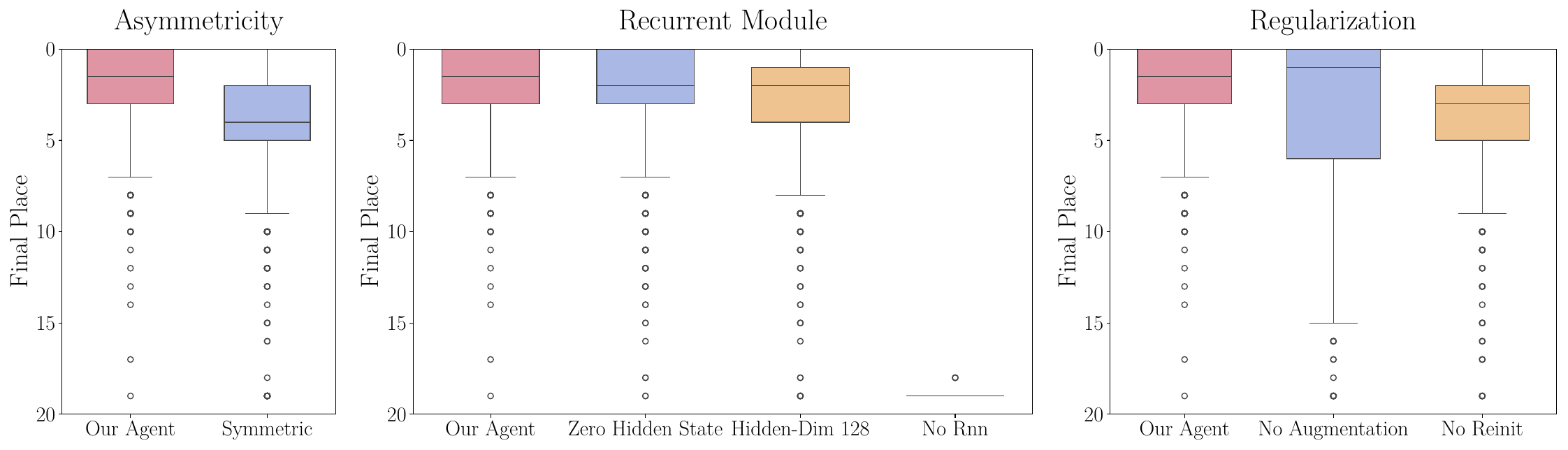} 
\vspace{2mm}
\caption{\textbf{Ablation studies on the Tokyo scenario.}
Ablated variations are compared by \textit{final place} metric, evaluated over 500 episodes with three random seeds.
\textbf{Left:} shows a comparison with the symmetric architecture variant;
%(c) Increasing the RNN sequence length does not result in performance gains while reducing it below our default of 32 impairs performance. Notably, a sequence length of 0 prevents the resulting agent from even overtaking a single opponent. 
\textbf{Middle:} compares variations with different recurrent module configurations;
\textbf{Right:} shows how each regularization approach affects performance. 
}
\label{fig:ablation}
\end{center}
\end{figure*}

\subsection{Visual Analysis}

To better understand the decision-making process of our vision-based agent, we apply Integrated Gradients (IG) \cite{sundararajan2017axiomatic}, which assigns importance to individual pixels in an image. In our analysis, attributions are calculated using a $7 \times 7$ mean-filtered blurred baseline as a reference point. Gradients are then computed over 20 linearly interpolated images between the baseline and the input image, integrating the resulting gradients along this path. These integrated attributions are summed to quantify the contribution of each pixel to the output action. Figure~\ref{fig:integrated_gradients} visualizes these attributions during a short driving segment, highlighting pixels contributing to the top 90\% of attribution values.
Past work has shown that pixels with high attributions are crucial to maintaining performance in the racing domain \cite{vasco2024super}.

The middle row of Figure~\ref{fig:integrated_gradients} shows that the agent exhibits context-dependent attention patterns. When near opponents, the agent focuses on lower vehicle regions and shadows to assess overtaking opportunities, similar to human drivers who rely on these cues in competitive driving \cite{nagy2024area}. On straight sections, the agent shifts attention to static features like vanishing points, treelines, and skylines, which aid in track localization and turn anticipation, reflecting the gaze patterns of professional drivers \cite{van2017differences}. While lane markings occasionally attract attention, their reliability is influenced by lateral positioning and the presence of opponents. As a result, the agent prioritizes more stable environmental cues, such as the sky and treelines.

The bottom row of Figure~\ref{fig:integrated_gradients} illustrates how the agent utilizes the recurrent module to capture long-term dependencies. By performing backpropagation-through-time \cite{werbos1990backpropagation} to compute IG, we visualize how information from earlier frames contributes to actions predicted for the final frame in the sequence. This visualization demonstrates how the agent uses early-frame data to infer opponent positions and trajectories, which informs its decision-making for future maneuvers. This ability to integrate long-term predictions is crucial for partially observable, multi-player environments like racing, where opponent velocities cannot be directly inferred from a single frame.

%\begin{figure*}
%\begin{subfigure}[b]{0.23\linewidth}
%    \includegraphics[width=\linewidth]{figures/Symmetric vs. %Asymmetric Architecture.pdf}
%    \caption{Asymmetricity}
%    \label{fig:symmetric_vs_asymmetric}
%\end{subfigure}
%\begin{subfigure}[b]{0.23\linewidth}
%    \includegraphics[width=\linewidth]{figures/Overfitting.pdf}
%    \caption{Regularization}
%    \label{fig:overfitting}
%\end{subfigure}
%\begin{subfigure}[b]{0.23\linewidth}
%    \includegraphics[width=\linewidth]{figures/RNN Sequence %Length.pdf}
%    \caption{RNN Sequence Length}
%    \label{fig:rnn_sequence_length}
%\end{subfigure}
%\begin{subfigure}[b]{0.23\linewidth}
%    \includegraphics[width=\linewidth]{figures/Hidden State Dims and Storing to Replay Buffer.pdf}
%    \caption{RNN Hidden State}
%    \label{fig:hidden_state_dims}
%\end{subfigure}

\subsection{Ablation Studies}

We conducted ablation studies to evaluate the contributions of specific architectural and training decisions to the performance of our vision-based agent. Each ablation involved modifications or removals of model components, followed by an evaluation to quantify their impact.  We trained agents in each setting for 5,000 epochs and selected the top-performing checkpoint from three seeds for each setting based on the highest \textit{winning margin}. Each model was evaluated over 500 episodes, and the results were summarized using box plots of each agent's \textit{final place}, the final place after the race.

%\begin{enumerate}[itemsep=0.5em, left=0.0em, label=\textbullet]
\begin{enumerate}

\item \textbf{Asymmetric Architecture:} The asymmetric architecture, where the critic incorporates both local (image and proprioceptive data) and global features, was pivotal for champion-level performance. In contrast, the symmetric variant, relying solely on local features, consistently failed to achieve first place in most evaluations. This result highlights the importance of leveraging global features in the critic.

\item \textbf{Recurrent Module:}
During training, initializing the recurrent module's hidden state to zeros before RNN warmup slightly degraded performance compared to using the replay buffer’s stored hidden state. Reducing the hidden state dimension from 512 to 128 caused a noticeable performance drop, while completely removing the RNN resulted in complete failure, with the agent unable to overtake any opponents. These findings emphasize the RNN's role in maintaining temporal continuity, tracking off-screen opponents, and estimating their velocity and direction.

\item \textbf{Regularization:} Applying image augmentation reduced the variance in performance across evaluation episodes, by potentially enhancing the agent's generalization and mitigating overfitting to specific visual inputs. Similarly, reinitializing the networks at 2000 epochs improved stability and final performance, as they allowed the agent to relearn and emphasize underrepresented features in the replay buffer, leading to a more balanced use of diverse visual inputs during training.

% \item \textbf{RNN Sequence Length:} 
% Integrating a recurrent architecture in the actor was crucial for performance. We tested sequence lengths of ${0, 16, 32, 64}$. When the sequence length was 0 (i.e., no RNN), the agent struggled to track opponents outside the visual frame or infer their velocity and direction, resulting in significant performance degradation. Notably, the agent without an RNN failed to overtake any opponents. In contrast, the RNN-equipped agent showed substantial improvement, with the best performance observed at a sequence length of 32. Larger sequence lengths (e.g., 64) led to diminishing returns, likely due to the high correlation between samples in the mini-batch.

% \item \textbf{Recurrent Module:} Increasing the RNN hidden dimension size improved performance, with 128 providing sufficient capacity for our task but resulting in slower learning. Initializing the replay buffer with all zeros during policy updates, rather than using the hidden state from data collection, negatively impacted performance. This was likely due to distributional shifts in the hidden state between training and collection, even with the default warmup of 16 steps.

\end{enumerate}

\section{CONCLUSION}

In this work, we introduced a vision-based autonomous racing agent that achieves champion-level performance in Gran Turismo 7, using only ego-centric camera views and proprioceptive data for inference. Our approach leverages an asymmetric actor-critic framework, where the actor uses both ego-centric and proprioceptive inputs, enhanced by a recurrent neural network to capture track layouts and opponent dynamics. The critic, on the other hand, has privileged access to the detailed track and opponent information during training. The agent consistently outperformed model predictive control drivers, achieving overtaking maneuvers comparable to, or better than, those of human champions. %Ablation studies highlighted the importance of image data augmentation and periodic network resets in optimizing performance.

This work sets a new benchmark for vision-based competitive racing and demonstrates the potential of reinforcement learning in high-performance, real-time environments. However, our agent was tested in a controlled setting, using a single vehicle type per scenario with fixed weather conditions. Future work could focus on expanding the framework to enable competitive head-to-head races with top human drivers, as well as enhancing the agent's ability to generalize across different car models, tracks, and weather conditions. Overcoming these challenges will be key to deploying vision-based racing agents in real-world scenarios.

%\addtolength{\textheight}{-12cm}   % This command serves to balance the column lengths
                                  % on the last page of the document manually. It shortens
                                  % the textheight of the last page by a suitable amount.
                                  % This command does not take effect until the next page
                                  % so it should come on the page before the last. Make
                                  % sure that you do not shorten the textheight too much.

%%%%%%%%%%%%%%%%%%%%%%%%%%%%%%%%%%%%%%%%%%%%%%%%%%%%%%%%%%%%%%%%%%%%%%%%%%%%%%%%

%%%%%%%%%%%%%%%%%%%%%%%%%%%%%%%%%%%%%%%%%%%%%%%%%%%%%%%%%%%%%%%%%%%%%%%%%%%%%%%%

%%%%%%%%%%%%%%%%%%%%%%%%%%%%%%%%%%%%%%%%%%%%%%%%%%%%%%%%%%%%%%%%%%%%%%%%%%%%%%%%

%Let's do double check to confirm if we can disclosure Mikail's name here.

\section*{ACKNOWLEDGMENTS}
We are very grateful to Polyphony Digital Inc. and Sony Interactive Entertainment for enabling this
research. We would also like to express our gratitude to Mikail Hizal, a GT champion, and Rodney Meza, an expert player in GT, for providing the reference data used in our evaluation.

%%%%%%%%%%%%%%%%%%%%%%%%%%%%%%%%%%%%%%%%%%%%%%%%%%%%%%%%%%%%%%%%%%%%%%%%%%%%%%%%

\bibliographystyle{IEEEtran}  
\bibliography{main}  

@article{fuchs2021super,
  title={Super-human performance in gran turismo sport using deep reinforcement learning},
  author={Fuchs, Florian and Song, Yunlong and Kaufmann, Elia and Scaramuzza, Davide and D{\"u}rr, Peter},
  journal={IEEE Robotics and Automation Letters},
  volume={6},
  number={3},
  pages={4257--4264},
  year={2021},
  publisher={IEEE}
}

@inproceedings{song2021autonomous,
  title={Autonomous overtaking in gran turismo sport using curriculum reinforcement learning},
  author={Song, Yunlong and Lin, HaoChih and Kaufmann, Elia and D{\"u}rr, Peter and Scaramuzza, Davide},
  booktitle={2021 IEEE international conference on robotics and automation (ICRA)},
  pages={9403--9409},
  year={2021},
  organization={IEEE}
}

@article{wurman2022outracing,
  title={Outracing champion Gran Turismo drivers with deep reinforcement learning},
  author={Wurman, Peter R and Barrett, Samuel and Kawamoto, Kenta and MacGlashan, James and Subramanian, Kaushik and Walsh, Thomas J and Capobianco, Roberto and Devlic, Alisa and Eckert, Franziska and Fuchs, Florian and others},
  journal={Nature},
  volume={602},
  number={7896},
  pages={223--228},
  year={2022},
  publisher={Nature Publishing Group UK London}
}

@article{vasco2024super,
  title={A Super-human Vision-based Reinforcement Learning Agent for Autonomous Racing in Gran Turismo},
  author={Vasco, Miguel and Seno, Takuma and Kawamoto, Kenta and Subramanian, Kaushik and Wurman, Peter R and Stone, Peter},
  journal={arXiv preprint arXiv:2406.12563},
  year={2024}
}

@article{cai2021vision,
  title={Vision-based autonomous car racing using deep imitative reinforcement learning},
  author={Cai, Peide and Wang, Hengli and Huang, Huaiyang and Liu, Yuxuan and Liu, Ming},
  journal={IEEE Robotics and Automation Letters},
  volume={6},
  number={4},
  pages={7262--7269},
  year={2021},
  publisher={IEEE}
}

@inproceedings{folkers2019controlling,
  title={Controlling an autonomous vehicle with deep reinforcement learning},
  author={Folkers, Andreas and Rick, Matthias and B{\"u}skens, Christof},
  booktitle={2019 IEEE Intelligent Vehicles Symposium (IV)},
  pages={2025--2031},
  year={2019},
  organization={IEEE}
}

@article{remonda2021formula,
  title={Formula rl: Deep reinforcement learning for autonomous racing using telemetry data},
  author={Remonda, Adrian and Krebs, Sarah and Veas, Eduardo and Luzhnica, Granit and Kern, Roman},
  journal={arXiv preprint arXiv:2104.11106},
  year={2021}
}

@inproceedings{herman2021learn,
  title={Learn-to-race: A multimodal control environment for autonomous racing},
  author={Herman, James and Francis, Jonathan and Ganju, Siddha and Chen, Bingqing and Koul, Anirudh and Gupta, Abhinav and Skabelkin, Alexey and Zhukov, Ivan and Kumskoy, Max and Nyberg, Eric},
  booktitle={proceedings of the IEEE/CVF International Conference on Computer Vision},
  pages={9793--9802},
  year={2021}
}

@inproceedings{haarnoja2018soft,
  title={Soft actor-critic: Off-policy maximum entropy deep reinforcement learning with a stochastic actor},
  author={Haarnoja, Tuomas and Zhou, Aurick and Abbeel, Pieter and Levine, Sergey},
  booktitle={International conference on machine learning},
  pages={1861--1870},
  year={2018},
  organization={PMLR}
}

@inproceedings{hafner2019learning,
  title={Learning latent dynamics for planning from pixels},
  author={Hafner, Danijar and Lillicrap, Timothy and Fischer, Ian and Villegas, Ruben and Ha, David and Lee, Honglak and Davidson, James},
  booktitle={International conference on machine learning},
  pages={2555--2565},
  year={2019},
  organization={PMLR}
}

@inproceedings{yarats2021improving,
  title={Improving sample efficiency in model-free reinforcement learning from images},
  author={Yarats, Denis and Zhang, Amy and Kostrikov, Ilya and Amos, Brandon and Pineau, Joelle and Fergus, Rob},
  booktitle={Proceedings of the aaai conference on artificial intelligence},
  volume={35},
  number={12},
  pages={10674--10681},
  year={2021}
}

@article{lee2024plastic,
  title={Plastic: Improving input and label plasticity for sample efficient reinforcement learning},
  author={Lee, Hojoon and Cho, Hanseul and Kim, Hyunseung and Gwak, Daehoon and Kim, Joonkee and Choo, Jaegul and Yun, Se-Young and Yun, Chulhee},
  journal={Advances in Neural Information Processing Systems},
  volume={36},
  year={2024}
}

@article{kostrikov2020image,
  title={Image augmentation is all you need: Regularizing deep reinforcement learning from pixels},
  author={Kostrikov, Ilya and Yarats, Denis and Fergus, Rob},
  journal={arXiv preprint arXiv:2004.13649},
  year={2020}
}

@article{yarats2021mastering,
  title={Mastering visual continuous control: Improved data-augmented reinforcement learning},
  author={Yarats, Denis and Fergus, Rob and Lazaric, Alessandro and Pinto, Lerrel},
  journal={arXiv preprint arXiv:2107.09645},
  year={2021}
}

@inproceedings{nikishin2022primacy,
  title={The primacy bias in deep reinforcement learning},
  author={Nikishin, Evgenii and Schwarzer, Max and D’Oro, Pierluca and Bacon, Pierre-Luc and Courville, Aaron},
  booktitle={International conference on machine learning},
  pages={16828--16847},
  year={2022},
  organization={PMLR}
}

@inproceedings{kapturowski2018r2d2,
  title={Recurrent experience replay in distributed reinforcement learning},
  author={Kapturowski, Steven and Ostrovski, Georg and Quan, John and Munos, Remi and Dabney, Will},
  booktitle={International conference on learning representations},
  year={2018}
}

@inproceedings{dabney2018distributional,
  title={Distributional reinforcement learning with quantile regression},
  author={Dabney, Will and Rowland, Mark and Bellemare, Marc and Munos, R{\'e}mi},
  booktitle={Proceedings of the AAAI conference on artificial intelligence},
  volume={32},
  number={1},
  year={2018}
}

@article{pinto2017asymmetric,
  title={Asymmetric actor critic for image-based robot learning},
  author={Pinto, Lerrel and Andrychowicz, Marcin and Welinder, Peter and Zaremba, Wojciech and Abbeel, Pieter},
  journal={arXiv preprint arXiv:1710.06542},
  year={2017}
}

@article{laskin2020rad,
  title={Reinforcement learning with augmented data},
  author={Laskin, Misha and Lee, Kimin and Stooke, Adam and Pinto, Lerrel and Abbeel, Pieter and Srinivas, Aravind},
  journal={Advances in neural information processing systems},
  volume={33},
  pages={19884--19895},
  year={2020}
}

@article{betz2022autonomous,
  title={Autonomous vehicles on the edge: A survey on autonomous vehicle racing},
  author={Betz, Johannes and Zheng, Hongrui and Liniger, Alexander and Rosolia, Ugo and Karle, Phillip and Behl, Madhur and Krovi, Venkat and Mangharam, Rahul},
  journal={IEEE Open Journal of Intelligent Transportation Systems},
  volume={3},
  pages={458--488},
  year={2022},
  publisher={IEEE}
}

@inproceedings{herrmann2020minimum,
  title={Minimum race-time planning-strategy for an autonomous electric racecar},
  author={Herrmann, Thomas and Passigato, Francesco and Betz, Johannes and Lienkamp, Markus},
  booktitle={2020 IEEE 23rd International Conference on Intelligent Transportation Systems (ITSC)},
  pages={1--6},
  year={2020},
  organization={IEEE}
}

@inproceedings{vazquez2020optimization,
  title={Optimization-based hierarchical motion planning for autonomous racing},
  author={V{\'a}zquez, Jos{\'e} L and Br{\"u}hlmeier, Marius and Liniger, Alexander and Rupenyan, Alisa and Lygeros, John},
  booktitle={2020 IEEE/RSJ international conference on intelligent robots and systems (IROS)},
  pages={2397--2403},
  year={2020},
  organization={IEEE}
}

@article{massa2020lidar,
  title={Lidar-based gnss denied localization for autonomous racing cars},
  author={Massa, Federico and Bonamini, Luca and Settimi, Alessandro and Pallottino, Lucia and Caporale, Danilo},
  journal={Sensors},
  volume={20},
  number={14},
  pages={3992},
  year={2020},
  publisher={MDPI}
}

@article{peng2021vehicle,
  title={Vehicle odometry with camera-lidar-IMU information fusion and factor-graph optimization},
  author={Peng, Wen-zheng and Ao, Yin-hui and He, Jing-hui and Wang, Peng-fei},
  journal={Journal of Intelligent \& Robotic Systems},
  volume={101},
  pages={1--13},
  year={2021},
  publisher={Springer}
}

@article{karle2022scenario,
  title={Scenario understanding and motion prediction for autonomous vehicles—review and comparison},
  author={Karle, Phillip and Geisslinger, Maximilian and Betz, Johannes and Lienkamp, Markus},
  journal={IEEE Transactions on Intelligent Transportation Systems},
  volume={23},
  number={10},
  pages={16962--16982},
  year={2022},
  publisher={IEEE}
}

@inproceedings{williams2018robust,
  title={Robust Sampling Based Model Predictive Control with Sparse Objective Information.},
  author={Williams, Grady and Goldfain, Brian and Drews, Paul and Saigol, Kamil and Rehg, James M and Theodorou, Evangelos A},
  booktitle={Robotics: Science and Systems},
  volume={14},
  pages={2018},
  year={2018}
}

@article{hao2022outracing,
  title={Outracing human racers with model-based autonomous racing},
  author={Hao, Ce and Tang, Chen and Bergkvist, Eric and Weaver, Catherine and Sun, Liting and Zhan, Wei and Tomizuka, Masayoshi},
  journal={arXiv preprint arXiv:2211.09378},
  year={2022}
}

@inproceedings{kalaria2024adaptive,
  title={Adaptive planning and control with time-varying tire models for autonomous racing using extreme learning machine},
  author={Kalaria, Dvij and Lin, Qin and Dolan, John M},
  booktitle={2024 IEEE International Conference on Robotics and Automation (ICRA)},
  pages={10443--10449},
  year={2024},
  organization={IEEE}
}

@inproceedings{xue2024learning,
  title={Learning model predictive control with error dynamics regression for autonomous racing},
  author={Xue, Haoru and Zhu, Edward L and Dolan, John M and Borrelli, Francesco},
  booktitle={2024 IEEE International Conference on Robotics and Automation (ICRA)},
  pages={13250--13256},
  year={2024},
  organization={IEEE}
}

@article{kazerouni2022slam_survey,
  title={A survey of state-of-the-art on visual SLAM},
  author={Kazerouni, Iman Abaspur and Fitzgerald, Luke and Dooly, Gerard and Toal, Daniel},
  journal={Expert Systems with Applications},
  volume={205},
  pages={117734},
  year={2022},
  publisher={Elsevier}
}

@article{macario2022slam_comprehensive,
  title={A comprehensive survey of visual slam algorithms},
  author={Macario Barros, Andr{\'e}a and Michel, Maugan and Moline, Yoann and Corre, Gwenol{\'e} and Carrel, Fr{\'e}d{\'e}rick},
  journal={Robotics},
  volume={11},
  number={1},
  pages={24},
  year={2022},
  publisher={MDPI}
}

@inproceedings{jaritz2018end,
  title={End-to-end race driving with deep reinforcement learning},
  author={Jaritz, Maximilian and De Charette, Raoul and Toromanoff, Marin and Perot, Etienne and Nashashibi, Fawzi},
  booktitle={2018 IEEE international conference on robotics and automation (ICRA)},
  pages={2070--2075},
  year={2018},
  organization={IEEE}
}

@inproceedings{raji2022motion,
  title={Motion planning and control for multi vehicle autonomous racing at high speeds},
  author={Raji, Ayoub and Liniger, Alexander and Giove, Andrea and Toschi, Alessandro and Musiu, Nicola and Morra, Daniele and Verucchi, Micaela and Caporale, Danilo and Bertogna, Marko},
  booktitle={2022 IEEE 25th International Conference on Intelligent Transportation Systems (ITSC)},
  pages={2775--2782},
  year={2022},
  organization={IEEE}
}

@article{remonda2024simulation,
  title={A Simulation Benchmark for Autonomous Racing with Large-Scale Human Data},
  author={Remonda, Adrian and Hansen, Nicklas and Raji, Ayoub and Musiu, Nicola and Bertogna, Marko and Veas, Eduardo and Wang, Xiaolong},
  journal={arXiv preprint arXiv:2407.16680},
  year={2024}
}

@inproceedings{todorov2012mujoco,
  title={Mujoco: A physics engine for model-based control},
  author={Todorov, Emanuel and Erez, Tom and Tassa, Yuval},
  booktitle={2012 IEEE/RSJ international conference on intelligent robots and systems},
  pages={5026--5033},
  year={2012},
  organization={IEEE}
}

@inproceedings{sundararajan2017axiomatic,
  title={Axiomatic attribution for deep networks},
  author={Sundararajan, Mukund and Taly, Ankur and Yan, Qiqi},
  booktitle={International conference on machine learning},
  pages={3319--3328},
  year={2017},
  organization={PMLR}
}

@article{nagy2024area,
  title={Area of Interest Tracking Techniques for Driving Scenarios Focusing on Visual Distraction Detection},
  author={Nagy, Viktor and F{\"o}ldesi, P{\'e}ter and Istenes, Gy{\"o}rgy},
  journal={Applied Sciences},
  volume={14},
  number={9},
  pages={3838},
  year={2024},
  publisher={MDPI}
}

@article{van2017differences,
  title={Differences between racing and non-racing drivers: A simulator study using eye-tracking},
  author={Van Leeuwen, Peter M and De Groot, Stefan and Happee, Riender and De Winter, Joost CF},
  journal={PLoS one},
  volume={12},
  number={11},
  pages={e0186871},
  year={2017},
  publisher={Public Library of Science San Francisco, CA USA}
}

@book{Scott1992,
  title={Multivariate Density Estimation: Theory, Practice, and Visualization},
  author={Scott, David W.},
  year={1992},
  publisher={Wiley-Interscience}
}

@article{cho2020learning,
  title={Learning phrase representations using RNN encoder-decoder for statistical machine translation. arXiv 2014},
  author={Cho, Kyunghyun and Van Merrienboer, Bart and Gulcehre, Caglar and Bahdanau, Dzmitry and Bougares, Fethi and Schwenk, Holger and Bengio, Yoshua},
  journal={arXiv preprint arXiv:1406.1078},
  year={2020}
}

@article{werbos1990backpropagation,
  title={Backpropagation through time: what it does and how to do it},
  author={Werbos, Paul J},
  journal={Proceedings of the IEEE},
  volume={78},
  number={10},
  pages={1550--1560},
  year={1990},
  publisher={IEEE}
}

@article{xiao2024anycar,
  title={AnyCar to Anywhere: Learning Universal Dynamics Model for Agile and Adaptive Mobility},
  author={Xiao, Wenli and Xue, Haoru and Tao, Tony and Kalaria, Dvij and Dolan, John M and Shi, Guanya},
  journal={arXiv preprint arXiv:2409.15783},
  year={2024}
}

\end{document}